\documentclass[manuscript,screen]{acmart}

\setcopyright{none}
\renewcommand\footnotetextcopyrightpermission[1]{}
\acmConference[FAccT '27]{ACM Conference on Fairness, Accountability, and Transparency}{June 2027}{}

\usepackage[english]{babel}
\usepackage{booktabs,multirow,makecell,array,tabularx}
\usepackage{threeparttable,adjustbox,float}
\usepackage{siunitx}
\usepackage{graphicx,subcaption,xspace}
\usepackage{listings,enumitem}
\usepackage[title]{appendix}
\definecolor{codegreen}{rgb}{0,0.6,0}
\definecolor{codegray}{rgb}{0.5,0.5,0.5}
\definecolor{codepurple}{rgb}{0.58,0,0.82}
\definecolor{backcolour}{rgb}{0.95,0.95,0.92}
\lstdefinestyle{prompt}{backgroundcolor=\color{backcolour},commentstyle=\color{codegreen},
  keywordstyle=\color{magenta},numberstyle=\tiny\color{codegray},stringstyle=\color{codepurple},
  basicstyle=\ttfamily\footnotesize,breaklines=true,captionpos=b,keepspaces=true,
  numbers=left,numbersep=5pt,showspaces=false,showstringspaces=false,tabsize=2}
\begin{document}

\title{Automated Call for Action: RAG-Based Environmental Law Enforcement Screening System}

\author{Matas Jurkevicius}
\email{matas.jurkevicius@student.uva.nl}
\correspondingauthor
\author{Yen-Chia Hsu}
\email{y.c.hsu@uva.nl}
\affiliation{%
  \institution{University of Amsterdam}
  \city{Amsterdam}
  \country{Netherlands}
}

\begin{abstract}
Environmental law inspectors face a structural issue in identifying potential violations from data: they often do not have the capacity to deal with large volume
of data present in the emission events. This paper examines to what extent an LLM-based system can support evidence assembly and permit-based screening of such events, using the coke plants of a steel producer and a regional environmental service as a case. A simple puff-trajectory model aligns air quality measurements, weather and citizen complaints with each emission event. A retrieval-augmented generation (RAG) pipeline then retrieves facility-specific permit conditions and generates event reports and a period report with an investigation recommendation. Across 74 events from March 2025, plant-aware re-ranking reduced reports citing the other plant’s conditions from 58.1\% (hybrid retrieval) to 0\% by rule-based plant labels, while maintaining high RAGAS and LLM-judge metrics. GPT-5-mini performed comparably to GPT-5.1, whereas GPT-5-nano scored lower on legal grounding and legal faithfulness. In an evidence ablation, removing air quality (AQ) data changed judge scores and period recommendations most; citation overlap changed only slightly under any removal, and removing complaint details had no measurable effect. This suggests a need for rules and guidelines that link complaints to incidents and permit conditions, so that community reports can count as evidence. The results support screening of events that may need further investigation. Design implications for involving environmental service experts are derived.
\end{abstract}


\maketitle

\section{Introduction}
\label{sec:introduction}

Dutch environmental services supervise whether residents and companies comply
with environmental law, grant environmental permits and check whether
permit holders meet their legal obligations \cite{odnzkg2026overons}. When an
operator reports an unusual incident or residents complain, inspectors must
relate the report to measurements, the circumstances of the incident and the
permit conditions involved \cite{gsnh2023vth}. Because enforcement decisions
must withstand objection and appeal \cite{gsnh2025bezwaar, gsnh2026intrekking},
they need an evidential basis; the precautionary principle alone, for
instance, cannot justify additional permit conditions \cite{gsnh2023vth}. This
work is carried out with limited capacity and specialist knowledge
\cite{vanaartsen2021}, while delays prolong residents' exposure to emissions
that affect their health (\S\ref{sec:case}). Tools that help inspectors
assemble and assess evidence are therefore needed.

Environmental services have begun experimenting with AI to detect such problems faster \cite{odnzkg2024jaarbeeld}. Automated tools, however, also influence which cases receive attention. Benami et al.\ \cite{benami2021} showed that in a US Clean Water Act programme, both the scope of the programme and design choices in a simulated risk model shifted enforcement attention between communities. Large Language Models (LLMs) bring additional risks to legal work. They perform well on many general tasks, but their performance on legal tasks varies widely with the type of legal reasoning required \cite{guha2023legalbench}, and commercial legal research tools built on retrieval still produce incorrect or unsupported citations \cite{magesh2024}. Both findings concern US law in English; to our knowledge, whether they carry over to Dutch environmental permits, which combine legal and technical Dutch, has not been tested.

This paper therefore builds a prototype screening system on publicly available
data of two Dutch coke and gas plants. Each incident that a factory notifies to
the environmental service, called an \emph{event}, is aligned with air quality
(AQ), weather and complaint data and written as an \emph{event paragraph}.
Permit conditions are retrieved for each event, and an LLM generates an
\emph{event report} that cites them. The event reports of a \emph{period} are
combined into a \emph{period report} that ends with an investigation
recommendation. The system is designed to support inspectors, not to replace
their judgement, and it supports screening, not compliance determination. The
study was carried out independently on public data only, and no public record
of which events inspectors considered worth investigating was found.
The evaluation therefore examines whether the system cites permit conditions
that apply to the emitting plant, stays faithful to its evidence and responds
to changes in that evidence, not whether its recommendations match the
decisions of the environmental service.

\subsection{Dutch Local Context about the Environmental Service and the Steel Industry}
\label{sec:case}
Our work is situated in a local context in the IJmuiden region in the Netherlands, where the OD NZKG (Dutch: Omgevingsdienst Noordzeekanaalgebied, the Dutch environmental service) supervises Tata Steel, one of the largest integrated steel production facilities in Europe \cite{buenopatin2026}, whose emissions significantly contribute to particulate matter, nitrogen dioxide, PAHs and metals in surrounding residential areas \cite{rivm2023tata}. Its coke and gas plants, KGF1 and KGF2, have been under enhanced supervision since 20 June 2023, following a decline in permit compliance, a substantial increase in unusual incidents and a third penalty payment order \cite{odnzkg2023}. Reducing air emissions from these incidents was one of the objectives set under enhanced supervision.
The events screened in this paper are the unusual incidents behind this enforcement history. Their notifications, the KGF permits and OD NZKG's evaluation reports are public, so the evidence for each event can be assembled without access to internal agency data.

\subsection{Research questions}
\label{sec:rqs}
This research aims to answer the main research question:
\textbf{to what extent can an LLM-based
screening system produce event and period reports that are grounded in the
permit conditions applicable to the emitting plant, faithful to the evidence,
and responsive to changes in that evidence?}
We aim to broaden the empirical knowledge about the capability of using LLMs in environmental permit reasoning. To support the answer and evaluate each stage of the compliance screening system, we have the following sub-research questions:

\begin{itemize}
\item \textbf{SRQ 1:} \textit{How can AQ, weather and complaint data be aligned in time and space with an event and written as an event paragraph?}
\item \textbf{SRQ 2:} \textit{Which retrieval strategies most effectively surface relevant permit conditions, and how do retrieval quality and the generator model affect the legal grounding of the event reports?}
\item \textbf{SRQ 3:} \textit{How do the event reports and the investigation recommendation of the period report change when one evidence stream (AQ, weather or complaints) is removed, and which stream has the largest effect?}
\end{itemize}
\section{Related Work}
\label{sec:related_work}

\subsection{Computational support for environmental enforcement}
Machine learning has been used to support environmental enforcement, for
example by predicting which facilities are likely to fail a water-pollution
inspection and proposing alternative inspection allocations
\cite{hino2018ml}. Such models rely on structured facility-level features, such
as permit type, facility type and exceedances of numeric permit limits
\cite{benami2021}. Their design choices, both policy and algorithmic, have
distributive consequences: in a US Clean Water Act compliance initiative, the
initial scope excluded facilities in denser areas with higher minority shares,
and the choice of prediction target modestly shifted attention between
communities \cite{benami2021}.

Systems built on reports raise a related concern, because reporting rates
differ between areas; in a predictive-policing simulation, differential
reporting displaced predicted hotspots towards high-reporting areas
\cite{Akpinar_2021}. Citizen complaints are one such report-based source. In a
national Chinese dataset, about a quarter of complaints could not be verified
on inspection, and complaint counts varied with per-capita income and internet
access \cite{ji2021citizen}. Complaints are therefore treated as context in
this paper, not as labels.

In contrast to these facility-level systems, this paper screens individual
events, with the text of permit conditions as the legal reference that an LLM
retrieves and cites. Which evidence streams a screening tool receives can
likewise shape what it reports; this is examined through an ablation of the
evidence streams (\S\ref{sec:res-ablation}).

\subsection{LLMs and retrieval for legal and regulatory text}
Benchmarks of legal reasoning show how LLMs perform on tasks close to compliance screening. LegalBench evaluates LLMs on 162 tasks covering six types of legal reasoning: issue-spotting, rule-recall, rule-application, rule-conclusion, interpretation and rhetorical understanding \cite{guha2023legalbench}. Rule-application is the type closest to deciding whether an emission event falls under a permit condition: models explain how a rule applies to a set of facts. Two differences matter here. In LegalBench the rule is supplied in the prompt, whereas screening an event first requires finding the relevant condition, and the benchmark is English-language and skewed towards American law and contracts.

Retrieval helps LLMs answer questions about long regulatory documents, but it
does not remove the risk of unsupported output. RAG conditions a generator on
passages retrieved from an external corpus so that outputs can be grounded in,
and traced to, source documents \cite{lewis2021, gao2024ragsurvey}. On
NEPAQuAD, a benchmark of question answering over US Environmental Impact
Statements, retrieved passages outperformed full documents; the benchmark uses
a single standard RAG configuration and does not compare retrieval variants
\cite{meyur2025}. In the legal domain, commercial research tools are built on
retrieval, yet they still hallucinated in roughly one in six to one in three of
the queries tested, although less often than a general-purpose model
\cite{magesh2024}.

Design choices in the retrieval stage strongly affect results. For legal
citation prediction on Australian case law, hybrid retrieval with a trained
re-ranker performed best among retrieval-based methods, and database
granularity and the choice of embedding model also mattered strongly
\cite{han2025}. The retrieval stage in this paper combines dense and lexical
rankings by weighted Reciprocal Rank Fusion \cite{cormack2009} and re-scores
candidates with an off-the-shelf cross-encoder
\cite{nogueira2020passagererankingbert}. Two features of this setting fall
outside the work above. Queries are English evidence summaries generated from
sensor and weather data, while the permit corpus is Dutch, so the dense encoder
must be multilingual \cite{wang2024me5}. In addition, one corpus regulates two plants whose conditions have overlapping vocabulary.

\subsection{Evaluation without ground truth}
Evaluating RAG outputs without annotated reference answers is a methodological challenge. RAGAS proposes reference-free metrics for RAG pipelines (faithfulness, answer relevance and context relevance) that can be computed without ground truth \cite{es2025ragas}; its software library, which we use, offers faithfulness, answer relevancy and a reference-free context precision metric. Faithfulness is measured against whatever was retrieved, so a report that cites a wrongly retrieved condition can still count as faithful, and none of these metrics assesses whether a report builds a coherent legal argument or acknowledges the limits of its evidence. LLM-as-a-judge evaluation scores free-form outputs with a strong model standing in for human raters. Zheng et al.\ found that GPT-4 agreed with human preferences on open-ended chat questions at a rate above 80\%, on a par with agreement between humans, but also documented position and verbosity biases and could not determine whether self-enhancement bias exists \cite{zheng2023}. This paper therefore combines three RAGAS metrics (faithfulness, answer relevancy and reference-free context precision) with a rubric-based LLM judge, and adds a plant-scope check that neither covers.

\subsection{Research gap}
Prior work thus covers parts of this setting: inspection-targeting models on
structured facility-level records \cite{hino2018ml, benami2021}, legal LLM
benchmarks and citation-retrieval studies on human-written questions and legal
passages \cite{guha2023legalbench, han2025}, and an environmental-review
benchmark on questions about documents \cite{meyur2025}. None of the identified work
screens individual incident notifications against permit conditions.
\section{Methodology}
\label{sec:methodology}

\subsection{Overview}
\label{sec:overview}

The methodology is structured in three stages. In stage 1, heterogeneous evidence streams are aligned into natural language event paragraphs. During stage 2, retrieval collects permit conditions requested by the query (event paragraph) and ranked by similarity. Finally, stage 3 generation creates evidence-supported event/period reports of the evidence using LLMs. The event generation step is evaluated using reference-free methods such as RAGAS and LLM-judge to measure quality and citation validity.

\subsection{Data}
\label{sec:data}

Six datasets are used.
\emph{Pollution events.}
Incident list filed by the factory and collected by the Dutch environmental service, including a timestamp, specific coke plant, and en event type (e.g., leakage or emission of a hazardous substance to air). This project uses the 2025 list, which has 707 events in total: 358 at coke plant KGF1 and 349 at KGF2.
\emph{Citizen complaints.}
Complaints filed by the citizens with the environmental service. Each complaint has a timestamp, a postcode area where it was filed from, a type and description of the event that the resident noticed.
A complaint is an unverified observation, not an official record. The 2025 data contain 813 complaints from 46 postcode areas.
\emph{Air quality.}
Hourly readings come from six air quality monitors of the national network of the Dutch National Institute for Public Health and the Environment (RIVM)~\cite{rivm_luchtmeetnet}.
The monitors are 0.3 to 4.2~km from the chimneys of the coke plants (station IDs in Appendix~\ref{app:dutch}).
Together they measure nine pollutants: black carbon (BC), CO, H$_2$S, NO, NO$_2$, NO$_x$, PM$_{10}$, PM$_{2.5}$ and SO$_2$.
Only two monitors measure gases; the other four measure PM$_{10}$ and PM$_{2.5}$.
\emph{Weather.}
Hourly observations of the Royal Netherlands Meteorological Institute (KNMI) at the Schiphol airport~\cite{knmi_uurgegevens} give wind direction and speed, temperature, solar radiation, cloud cover and humidity.
Other weather stations nearer to the plants have large gaps in these variables.
Schiphol has complete records but lies about 25~km from the plants.
\emph{Postcode areas.}
The 2024 PC4 boundaries of Statistics Netherlands (CBS)~\cite{cbs_pc4} let us place complaints on the plume cone (Section~\ref{sec:cone}).
\emph{Permit conditions.}
The environmental service agency publishes the permit conditions in Dutch of the two coke plants~\cite{odnzkg_tatasteel}.
The file has 50 records; we remove the 8 revoked records and keep 42 permit conditions, each with an identifier (for example 1.2.20.H.f), a text that describes the permit condition, a theme (e.g., Air and Odor), and a status (e.g., In force; Revoked).

We evaluate Stages~2 and~3 on March 2025 only, which has 74 events (27 at plant~1, 47 at plant~2) that can be reasonably verified within our limited computation budget. Furthermore, this month has the second highest number of events of the year and the most matched complaints (Appendix~\ref{app:months}).

\subsection{Evidence Aggregation}
\label{sec:aggregation}

For each event, Stage~1 finds the monitors and postcode areas that the emission can reach, scores the air quality data, matches complaints, and writes the event paragraph.

\subsubsection{Atmospheric Dispersion Model}
\label{sec:cone}

Each event is treated as a short release (a puff) that moves in a straight line with the wind from the chimney of the plant. This method estimates where an emission travels and when it arrives, not how concentrated the emission is, as the public data is not consistent and accurate enough to make correct calculations.

\emph{Wind and speed.}
The weather at the time of an event is estimated from the two neighbouring hourly observations by linear interpolation.
The plume travels opposite to the wind, so its direction is the wind direction plus 180$^\circ$, and when the wind is calm or variable (KNMI codes 0 and 990) at either neighbouring hour, the direction is undefined and the event has no monitor or postcode area in the plume.
Atmospheric stability is classified with the Pasquill-Gifford scheme~\cite{turner1994} (Appendix~\ref{app:margin}), and the wind speed is scaled to account for the chimney height of 35~m (as factory facility metrics are not public, an approximation based on general knowledge was used) with a power law~\cite{irwin1979}.
The puff moves at the calculated speed, with a minimum of 0.5~m/s.

\emph{Cone.}
A target (i.e., an air quality monitor or the centroid of a postcode area) lies in the plume cone if the angle from the chimney to the target differs from the plume direction by at most $\theta_s(x) + 15^\circ$, where $x$ is the distance in km and $s$ is the stability class, and
\begin{equation}
\theta_s(x) = c_s - d_s \ln x .
\end{equation}
Here $\theta_s(x)$ is the angle at which the concentration across the plume falls to 10\% of its centre value~\cite{turner1994,epa1995isc3v2}, which narrows with distance (constants in Appendix~\ref{app:margin}).
The 15$^\circ$ margin covers the uncertainty of the wind direction, which KNMI rounds to 10$^\circ$ and measures hourly about 25~km away, and it was calibrated on the 11 months other than March from the range in which AQ severity stays raised beyond the plume edge (Appendix~\ref{app:margin}).

\emph{Travel window.}
Once puff reaches an AQ monitor, readings are taken based on the approximated time the emissions should have reached it.
For a postcode area, the travel window starts at the time when the puff reaches the nearest point of the area and ends when it has passed the farthest point, after which a 90-minute window is given to account for delays in reporting. A width allowance of one plume width divided by the speed, used to account for uncertainty from chosen weather readings and include postcodes/monitors that might have been affected.
Travel window stops at 3 hours to prevents events that happened in extremely slow wind conditions from travelling indefinitely, as any reading matches become unlikely.

\subsubsection{Air Quality Scoring}
\label{sec:aqscore}

For each pollutant and monitor, a reading is expressed as a z-score, which is the reading minus the mean of the previous 30 days, divided by the standard deviation over the same 30 days,
\begin{equation}
z_t = \frac{x_t - \mu_t}{\sigma_t} .
\end{equation}
The window holds 720 hourly readings, including the $x_t$ reading itself, and a z-score is computed only with at least 72 readings (three days).
A PM$_{10}$ reading of 48.26~$\mu$g/m$^3$ against a mean of 24.3 and a standard deviation of 12.1, for example, gives $z = 1.98$.
The z-score shows how unusual a reading is for that monitor.
The 30-day window was chosen to separate seasonal trends from short-term anomalies and was not optimised; Section~\ref{sec:eval-srq1} tests how sensitive the ranks are to this choice.
Note that the time the puff reaches the monitor may not be at the exact time that we have the air quality measurement, so the z-score is then linearly interpolated between the two hourly values around the time the puff reaches the monitor.

The primary monitor of an event is the air quality monitor in the plume cone that is closest to the chimney.
If no monitor is in the cone, events at coke plant KGF2 use the monitor 0.3~km from it as a fallback since it is close enough to provide relevant information, while KGF1 has no nearby monitors and therefore has no fallbacks to prevent the lack of readings.

\subsubsection{Pollutant Classification}
\label{sec:classification}

To assess impact without obscuring pollutant-specific spikes, we evaluate air quality severity using separate groups for coke-specific and general emissions.
The coke pollutants (SO$_2$, H$_2$S, NO, NO$_x$) are products of coke oven gas combustion and of fugitive emissions~\cite{eubref2013,emepeea2023}, whereas the general pollutants (BC, PM$_{10}$, PM$_{2.5}$, CO) escape from combustion facilities.
NO$_2$ forms in the air by oxidation and is not a direct product of the ovens, so it is reported by itself.
The air quality severity of an event is the highest z-score of two pollutant groups at the primary monitor (defined based on \cite{epa2018}).
When events happen concurrently (more than one occur in the same hours), the air quality reading cannot be assigned to one of them, so for an event with $n$ other events (at either plant) within 3 hours before or after, the severity is divided equally:
\begin{equation}
\text{AQ severity} = \frac{1}{1+n}\,\max\left(z_{\text{coke}},\, z_{\text{general}}\right),
\end{equation}
where $z_{\text{coke}}$ and $z_{\text{general}}$ are the highest z-scores in the two groups.

\subsubsection{Complaint Matching}
\label{sec:complaints}

A citizen complaint matches an event if it comes from a postcode area whose centroid lies in the plume cone and its time falls within the travel window.
The event descriptions are grouped into coke-related (coke, sulphur or rotten eggs, bakelite or phenol, tar, ammonia), particulate (black or grey, dust cloud), and other (Appendix~\ref{app:dutch}). For each event paragraph, we aggregate the following information: number of matched complaints per descriptors and the range of their distances to the chimney.
Note that citizen complaints serve as context for the event when generating event paragraphs, not as labels of a violation.

\subsubsection{Event Paragraph Structure}
\label{sec:paragraph}

The event paragraph is built from a structured template that we designed and has four parts: the notification (timestamp, coke plant, pollution event type); weather (wind, stability class, plume direction, details whether  additional events occurred or if no postcode was crossed); air quality (the primary monitor, the z-score and reading of each pollutant, the two group z-scores, the air quality severity, and a disclaimer that readings are linearly interpolated, therefore not fully accurate); and aggregated information of citizen complaints (counts per group and distances, or the reason for not having the information).
The \emph{no-postcode flag} is set when no postcode area lies in the plume cone within 3 hours of travel, so that the plume goes over sea or unpopulated land.

\subsection{Permit Retrieval}
\label{sec:retrieval}

\subsubsection{Permit Corpus Construction}
\label{sec:corpus}

\emph{Permit condition} is a single chunk, consisting of an identifier and description of the permit, allowing each condition to be easily separated and cited.
While no information pre-processing was made, the embedding models were capable of reading at most 512 tokens per input, truncating permits that are too long.
Each condition is embedded with its identifier, and multilingual embedding models are used because the permits are Dutch and the query is English.
Since the conditions are not explicitly labelled, each condition was assigned a plant it applies to.
A condition belongs to coke plant KGF1 if its title or text names contain emission point numbered 1xx (EL or AV followed by three digits), 'KGF1' or 'Kooksgasfabriek 1', and to plant KGF2 if it names a 2xx emission point 'KGF2' or 'Kooksgasfabriek 2'; all other permit conditions are labelled as "shared" between two coke plants, including those that name both plants.
The rule uses text only, and the labels were confirmed manually.
A report for a specific coke plant that cites a permit condition labelled for the other plant is considered a \emph{wrong-plant citation}.

\subsubsection{Retrieval Architecture}
\label{sec:variants}

Four variants are compared (Table~\ref{tab:variants}).
Variant~A is a reference configuration (a small multilingual embedding model, dense retrieval only, three permit conditions) and is not part of the controlled comparison, whereas Variants~B, C and~D share the embedding model, number of permit conditions and add 1-2 components each to the one before.

\begin{table}[t]
\caption{Retrieval variants. $k$ is the number of conditions given to the generator. The pool is the number of conditions that the re-ranker scores.}
\label{tab:variants}
\small
\begin{tabular}{@{}lllllll@{}}
\toprule
Variant & Embedding model & BM25 & Reranker & Plant filter & Pool & $k$ \\
\midrule
A & multilingual-e5-base & no & no & no & -- & 3 \\
B & multilingual-e5-large & yes & no & no & -- & 8 \\
C & multilingual-e5-large & yes & yes & no & 16 & 8 \\
D & multilingual-e5-large & yes & yes & yes & 32 & 8 \\
\bottomrule
\end{tabular}
\end{table}

\emph{Dense retrieval and BM25.}
In Variant B, the event paragraph and the permit conditions are converted into embedding vectors using a larger multilingual-e5~\cite{wang2024me5}, ranked by cosine similarity, which are also ranked by BM25 (Okapi)~\cite{robertson2009bm25} on lowercase words separated by spaces.
Because the query is in English and the permit conditions are in Dutch, BM25 can match only shared tokens, such as emission point codes, pollutant symbols and the Dutch event types.
The two rankings are then combined by weighted Reciprocal Rank Fusion (RRF)~\cite{cormack2009},
\begin{equation}
\mathrm{RRF}(d) = \sum_{r \in \{\text{dense},\,\text{BM25}\}} \frac{w_r}{60 + \mathrm{rank}_r(d)},
\end{equation}
with $w_{\text{dense}} = 0.7$ and $w_{\text{BM25}} = 0.3$.
Variant~B returns conditions with top-8 highest RRF scores.

\emph{Reranking.}
Variant~C, expands on Variant~B with a cross-encoder re-ranking stage to improve contextual relevance. Rather than applying the cross-encoder to all permit conditions,  RRF is used to retrieve a candidate pool of the 16 highest-ranked conditions (amount chosen to test restrictive selection). The cross-encoder (ms-marco-MiniLM-L-6-v2)~\cite{nogueira2020passagererankingbert} then scores each event-paragraph and permit-condition pair. The final ranking combines the scaled cross-encoder and RRF scores:
\begin{equation}
s(d) = 0.8\,\tilde{c}(d) + 0.2\,\tilde{r}(d),
\end{equation}
where $\tilde{c}$ is the scaled cross-encoder score and $\tilde{r}$ the scaled RRF score.

\emph{Plant-aware re-ranking.}
Variant~D is identical to Variant~C, but it uses 32 permit conditions (amount increased to include more permit conditions for plant-aware re-ranking), as well as lower the priority of the permit conditions that are not related to the target coke plant through keywords when generating the report.
Conditions labelled for the wrong coke plant are moved to the bottom, while the larger initial pool leaves more candidates after the move.
If fewer than eight conditions are retrieved, conditions of the other plant fill the gaps in rank order, so wrong-plant citations remain possible.

\subsection{Report Generation}
\label{sec:generation}

\subsubsection{Prompt Design}
\label{sec:prompts}

The prompt for an event instructs the LLM to act as a compliance assessment assistant for the environmental services, to cite only permit conditions and evidence in the context, not to infer rules that were not retrieved, to clearly state whether a relevant condition is missing or not, and to write in English.
It supplies the name of the coke plant, the time, the event type, the event paragraph, the event data and the retrieved conditions (identifier, theme, status, and text), and asks for an event report with five sections: 1) event summary, 2) severity assessment, 3) permit conditions that may be relevant, 4) preliminary compliance observations, and 5) data gaps.

The prompt for a period report contains the period statistics (number of events, events with the no-postcode flag, events with reduced severity, mean and highest air quality severity) and all event reports with the identifiers of their retrieved conditions.
It asks for five parts: executive summary, recurring permit conditions, events ranked by air quality severity, the recommendation, and data limitations.
The recommendation states whether the period requires a formal investigation and begins with Yes, No or Inconclusive.
Appendix~\ref{app:prompts} gives prompts for generating events and period reports.

\subsubsection{LLM Generators}
\label{sec:generators}

Variants are tested on GPT-5.1. Two more LLM generators are added and compared to GPT-5.1 during model comparison: GPT-5-mini and GPT-5-nano.
Each LLM receives the same prompts, and each event report is generated once, using in-house infrastructure for GDPR compliance.

\subsection{Evaluation}
\label{sec:evaluation}

\subsubsection{Aggregation Statistical Evaluation}
\label{sec:eval-srq1}

Three checks are run on the event paragraphs (SRQ1).

\emph{Bounds and sanity checks.}
38 checks are run on all 707 events: weather values and plume directions must lie in physical ranges, concentrations must not be negative, and no z-score may exceed 10 in absolute value.
An event without an air quality severity must also have a reason, namely that no monitor was assigned or that the monitor has fewer than 72 earlier readings (the first three days of 2025).

\emph{Complaints and air quality severity checks.}
To understand if citizens indicate elevated severity, the air quality severity of events with at least one matched complaint is compared with that of events with none, using a two-sided Mann-Whitney U test on events that have an air quality severity.
Complaints and monitors are placed by the same plume cone and travel window, so correct alignment predicts higher air quality severity when complaints occur; a difference is consistent with correct alignment but does not prove it, because other factors, such as the coke plant and the choice of monitor, can also act as confounding factors that produce a difference.

\emph{Window length.}
For the March events, the highest z-score at the primary monitor at the event time is computed with windows of 7, 14, 30 and 60 days, and the ranks from the 7-, 14- and 60-day windows are compared with the 30-day ranks by Spearman's rank correlation.
A high correlation shows that the ranking is insensitive to the window; it does not show whether 30 days is the best choice.

\subsubsection{RAGAS Metrics}
\label{sec:eval-ragas}

Four scores of the RAGAS library~\cite{es2025ragas} are used, which need no reference answers, with gpt-oss-120b as evaluator (SRQ2). We briefly explain them below:

\emph{Faithfulness (Legal)} measures the share of claims made in sections 3 and 4 of the event report that are supported by the retrieved permit conditions. The evaluator first extracts individual claims from these sections and assesses whether each claim can be inferred from the permit conditions. The score is the number of supported claims divided by the total number of claims, ranging from 0 to 1, with higher values indicating better results.

\emph{Faithfulness (Factual)} measures the same information but for the entire report, using both the event paragraph and the retrieved permit conditions as supporting context. It uses the same calculation and score criteria, but also considers factual support from the event paragraph.

\emph{Answer Relevancy} measures how well the report addresses the event paragraph using multilingual-e5-base embeddings. The evaluator generates a question from the report, embeds it alongside the event paragraph, and calculates their cosine similarity. The score ranges from 0 to 1, with higher values indicating better results.

\emph{Context Precision} measures whether retrieved permit conditions relevant to the event paragraph are ranked above less relevant conditions. The evaluator first assesses the relevance of each retrieved condition and then calculates the average precision at the ranks of relevant conditions. The score ranges from 0 to 1, with higher values indicating better results. 

\emph{Context Recall} requires reference answers and is therefore not computed.

\subsubsection{LLM-as-a-judge Evaluation}
\label{sec:eval-judge}

The LLM judge (gpt-oss-120b) scores each event report from 1 to 3 on four dimensions~\cite{zheng2023}, listed and explained below.

\emph{Legal Grounding (LG)} evaluates whether every compliance claim in sections 3 and 4 cites a specific retrieved permit condition, whether the condition exists in the retrieved set, and whether it applies to the correct chimney.

\emph{Evidence Integration (EI)} assesses whether AQ sensor readings, atmospheric conditions (wind and PG class), and citizen reports are connected to the legal reasoning in sections 3 and 4, rather than merely listed in section 2.

\emph{Epistemic Calibration (EC)} measures how well the report notes the limitation that ambient readings in $\mu$g/m$^3$ cannot be compared with stack limits in mg/Nm$^3$ and avoids claiming compliance or violation from ambient data.

\emph{Data Gap Identification (DG)} evaluates whether section 5 identifies specific evidence needed for a complete LLM assessment, including evidence relevant to the retrieved permit conditions. 

The maximum score is thus 12 (Appendix~\ref{app:rubric}). For each dimension, 1 means failure to account for the information, 2 means sufficient but incomplete disclosure of relevant information, 3 means complete explanation.

\emph{Wrong-Plant Citation Rate (WP\%).} The judge also flags a report that cites a condition belonging only to the other coke plant, using the rule in Section~\ref{sec:corpus}, and a report counts as a wrong-plant report if at least two of the three runs flag it.
The \emph{wrong-plant rate}, the share of events with a wrong-plant report, is the main metric for comparing retrieval variants, because RAGAS Context Precision cannot evaluate which permit condition belongs to which coke plant.

For the main evaluation, the judge reads the event paragraph, the full text of the retrieved conditions and the report, and each report is scored in three runs whose scores are then averaged.

\subsubsection{Evidence Ablation}
\label{sec:eval-ablation}

In each ablation cycle, one evidence stream is removed from the event paragraphs, and the event/period reports are generated again.
The four cycles tested in this experimentation are as follows: baseline (nothing removed), no AQ (removed monitor readings, z-scores and the AQ attribution of concurrent events), no weather (removed wind, stability class, plume direction and the statements connected to them) and no complaints (removed complaint counts, descriptors, distances and postcodes).
Removed data is replaced by a text statement (e.g., ``[Air quality data not available for this event.]''), while other readings are pieced together into a new event paragraph.
 
Each condition is generated three times for all 74 events with GPT-5.1. Generation uses temperature 1.0 so that repeated runs vary and the noise floor can be measured (Table~\ref{tab:ablation}). The retrieved conditions are the same in every condition, so the removal changes only the evidence that the generator receives.

Four measures are reported for each setting.

\emph{Overlap} is the Jaccard index of the permit
conditions cited in two reports for the same event: the size of the shared set
divided by the size of the combined set. For the baseline row, overlap is
computed between the three baseline repeats and averaged over events; this
value is the noise floor, i.e., the variation without any removal. For a
removal setting, overlap is computed between each repeat of that setting and
the baseline repeat of the same event and averaged over events. A setting
whose overlap is close to the noise floor changes the cited conditions no more
than repetition alone does.

\emph{Recommendation} is the first word of the period report (Yes, No or
Inconclusive). The majority over the three repeated runs of a setting is
reported.

\emph{Judge total} is the sum of the four LLM-judge dimensions
(Section~\ref{sec:eval-judge}; maximum 12), averaged over the 222 reports of a
setting (74 events $\times$ 3 repeats) and reported as the difference to the
baseline. \emph{EI} and \emph{EC} are the mean scores (1--3) of the evidence
integration and epistemic calibration dimensions. For ablation, each report is scored once by
gpt-oss-120b, which reads the event and is not blinded to the setting. The
judge does not assess whether a report is correct, only whether the report uses
the evidence it was given.
\section{Results}
\label{sec:results}
 
Stage~1 results cover all 707 events of 2025, and Stage~2 and Stage~3 results cover the 74 events of March~2025.
 
\subsection{Evidence Aggregation (SRQ1)}
\label{sec:res-srq1}

All 38 bounds and sanity checks on the 707 event paragraphs passed.
Air quality severity is missing for 120 events (17.0\%): 117 coke plant KGF1 events have no air quality monitor in the plume cone, and three events fall on 2 and 3 January, when the monitor had fewer than 72 earlier readings.
Of the 590 events with a primary monitor, 430 have it in the plume cone and 160 use the fallback monitor, and the wind direction is undefined for 22 events.
Coke pollutants are available for 129 events (18.2\%) and PM$_{10}$ for 582 (82.3\%), because only two of the six monitors measure gases.

Events with at least one matched citizen complaint have a higher air quality severity than events without (median 0.59 for 52 events against 0.26 for 535; two-sided Mann-Whitney $p = 0.006$).
The direction is the same at both coke plants, but the difference is significant only at plant KGF2 ($p = 0.012$, 29 against 317 events; plant~1: $p = 0.22$, 23 against 218 events), so the result is consistent with a correct alignment without proving it.

Replacing the 30-day window by 14 or 60 days changes the ranking of air quality severity little (Spearman $\rho$ = 0.96 and 0.99 on the 64 March events with a primary monitor), and the 7-day window changes it more ($\rho$ = 0.86).

\subsection{Retrieval Variants (SRQ2)}
\label{sec:res-retrieval}
 
Plant-aware re-ranking separate the variants more clearly than any other measure. As shown in the WP\% column in Table~\ref{tab:retrieval}, Variant~D reports contain 0 wrong-plant citations, followed by Variant~A, Variant~C, and finally Variant~B. Additionally, Variant~D achieves highest faithfulness, legal grounding, evidence integration and overall judge scores.

Citation diversity also differs substantially between variants. Variant A cites only six distinct conditions across the 74 events, compared with 17–21 for Variants B–D, yet achieves competitive evaluation scores. This indicates that broader citation diversity does not necessarily translate into better report quality.
 
\begin{table*}[t]
\caption{Retrieval variants (using generator GPT-5.1), 74 events in March~2025. RAGAS: F$_\text{leg}$ faithfulness (legal), F$_\text{fact}$ faithfulness (factual), Rel.\ answer relevancy, Prec.\ context precision. Judge (1--3 per dimension, total out of 12, mean of three runs): LG legal grounding, EI evidence integration; WP\% is the share of events with a wrong-plant report (majority of three runs, lower is better). Epistemic calibration (2.89--2.95) and data gap identification (3.00) do not differ between variants and are omitted.}
\label{tab:retrieval}
\centering
\small
\begin{tabular}{@{}lrrrrrrrr@{}}
\toprule
& \multicolumn{4}{c}{RAGAS} & \multicolumn{4}{c}{Judge} \\
\cmidrule(lr){2-5}\cmidrule(lr){6-9}
Variant & F$_\text{leg}$ & F$_\text{fact}$ & Rel. & Prec. & Total & LG & EI & WP\% \\
\midrule
A (dense, $k$=3)        & 0.542 & 0.806 & 0.852 & 0.998 & 10.79 & 2.75 & 2.12 & 29.7 \\
B (hybrid)              & 0.577 & 0.816 & 0.796 & 0.955 & 10.55 & 2.47 & 2.18 & 58.1 \\
C (hybrid + reranker)   & 0.591 & 0.791 & 0.784 & 0.931 & 10.78 & 2.61 & 2.23 & 47.3 \\
D (plant filter)        & 0.653 & 0.827 & 0.844 & 0.945 & 11.29 & 2.98 & 2.40 & 0.0 \\
\bottomrule
\end{tabular}
\end{table*}

\subsection{LLM Generators (SRQ2)}
\label{sec:res-generators}

A further investigation in Variant~D (see Table~\ref{tab:generators}) using the GPT-5 family with various model complexity shows that the performance of GPT-5-mini is as comparative as the more complex model GPT-5.1, while GPT-5-nano performs worse, especially in the faithfulness (legal), legal grounding and wrong-plant citation scores.
 
\begin{table*}[t]
\caption{LLM Generators with Variant~D retrieval, 74 events in March~2025. Columns as in Table~\ref{tab:retrieval}, with EC epistemic calibration. Data gap identification is 3.00 for all generators and is omitted. The Variant~D row of Table~\ref{tab:retrieval} scores the same reports as the GPT-5.1 row; the differences between the two rows indicate the noise of the scores (judge total 0.14, RAGAS up to 0.03).}
\label{tab:generators}
\centering
\small
\begin{tabular}{@{}lrrrrrrrrr@{}}
\toprule
& \multicolumn{4}{c}{RAGAS} & \multicolumn{5}{c}{Judge} \\
\cmidrule(lr){2-5}\cmidrule(lr){6-10}
Generator & F$_\text{leg}$ & F$_\text{fact}$ & Rel. & Prec. & Total & LG & EI & EC & WP\% \\
\midrule
GPT-5.1     & 0.644 & 0.800 & 0.843 & 0.941 & 11.15 & 2.95 & 2.29 & 2.91 & 0.0 \\
GPT-5-mini  & 0.638 & 0.778 & 0.857 & 0.951 & 11.35 & 2.97 & 2.47 & 2.91 & 1.4 \\
GPT-5-nano  & 0.478 & 0.731 & 0.824 & 0.907 & 10.82 & 2.77 & 2.37 & 2.67 & 4.1 \\
\bottomrule
\end{tabular}
\end{table*}
 
\subsection{Evidence Ablation (SRQ3)}
\label{sec:res-ablation}

Overlap changed only slightly under every removal (0.81–0.82 against 0.84) (Table~\ref{tab:ablation}). Period recommendations shift to ``Yes'' in two of three repeats without air quality data and one of three repeats without weather, while all other reports remain ``Inconclusive''. Removing air quality data produces the largest change in judge scores, particularly in evidence integration and epistemic calibration, suggesting the importance of AQ details when generating reports. Without weather, judge scores increase slightly, although the improvement is modest. Removing citizen complaints produces virtually no change in judge scores.

\begin{table*}[t]
\caption {Evidence removal, 74 events in March~2025 using GPT-5.1, three repeated runs per ablation cycle. Overlap is the mean Jaccard index of the cited conditions. Recommendation: majority of period recommendation that repeated runs suggest (Yes/No/Inconclusive). Judge: mean total out of 12 over 222 reports per condition (difference to the baseline), and mean evidence integration (EI) and epistemic calibration (EC) on the scale 1 to 3.}
\label{tab:ablation}
\centering
\small
\begin{tabular}{@{}lcccrr@{}}
\toprule
Experiment Setting & Overlap & Recommendation & Judge total (difference) & EI & EC \\
\midrule
Baseline      & 0.84 & Inconclusive & 11.27          & 2.39 & 2.92 \\
No AQ         & 0.81 & Yes          & 9.89 ($-1.38$)  & 1.66 & 2.26 \\
No weather    & 0.81 & Inconclusive & 11.54 ($+0.28$) & 2.70 & 2.88 \\
No complaints & 0.82 & Inconclusive & 11.26 ($-0.01$) & 2.37 & 2.93 \\
\bottomrule
\end{tabular}
\end{table*}
\section{Discussion}
\label{sec:discussion}

A prototype was built and evaluated on one month of events, indicating the feasibility of using LLMs to support environmental law enforcement, as shown in the result of Variant~D in Table~\ref{tab:retrieval}.
We now discuss each step in the pipeline.


\subsection{Evidence Aggregation}

The results show that evidence aggregation in general performs and connects information correctly.
The event paragraphs pass all consistency checks (none of the 38 tests found a violation) and agree with the complaints in the expected direction: events with matched complaints have a higher AQ severity than events without, as a correct alignment of plume, monitors and postcode areas predicts, but the difference is significant only at coke plant KGF2 (Section~\ref{sec:res-srq1}), which could be influenced by this coke plant having a fallback monitor and therefore more events with readings than KGF1.

Note that the evidence stays at the level of ambient air, whereas the permit conditions regulate the source.
This is a design constraint, as public air quality data from the emission source (the coke plant) cannot be obtained prior to a formal inspection. Initiating such an inspection requires identifying an unusual pollution event, which first requires analysing existing public air quality data.
Furthermore, only two of the six air quality monitors measure gases, so coke pollutants are available for 18.2\% of the events, no air quality severity exists for 17.0\% of the events, and the substances under enforcement (PAHs, benzene, benzo(a)pyrene and heavy metals) are not measured.
The event paragraphs therefore show whether a plume plausibly reached residents, but not whether a stack limit was exceeded, because the permit regulates the chemicals from the emission source and not the ambient air.


\subsection{Permit Retrieval}

A permit that covers several coke plants makes the retrieval step return the wrong plant's permit conditions.
When the permit conditions of both coke plants are in one set, the hybrid retrieval approach cited a condition of the other plant in 58.1\% of the event reports, the re-ranking approach reduced this number to 47.3\%, and the plant filter removed it (Table~\ref{tab:retrieval}).
Variants~B and~C retrieved on average 2.1 and 2.8 conditions of the other plant per event, out of eight. The likely, untested reason for the higher number in Variant~C is that a cross-encoder trained on English computes relevance based on topics and not the coke plant.
For similar work, conditions can be labelled by scope in advance, which can be used as a filter before ranking.

One point to note is that the 0.0\% wrong-plant rate of Variant~D shows that the plant-aware prioritisation operates as intended, not that it selects the correct permit conditions. The plant labels on which both the prioritisation and the evaluation rely were manually identified and assigned by the authors.

Two RAGAS scores, faithfulness (factual) (0.79 to 0.83) and context precision (0.93 to 1.00), hardly separate the variants, although the wrong-plant rate ranges from 0.0\% to 58.1\%.
Context precision is highest for Variant~A (0.998) because it retrieves only three conditions, and it cannot see which plant a condition belongs to.


\subsection{Report Generation}

The result shows that a smaller model writes event reports of similar quality (see Section~\ref{sec:results} and Table~\ref{tab:generators}).
Additionally, generated period reports indicate that different LLMs can give similar period reports but different investigation recommendations.
GPT-5.1, GPT-5-mini and GPT-5-nano all state that stack emission data and relevant legal documents are absent. However, GPT-5-mini concludes ``Yes'' (an investigation is needed to obtain the records), GPT-5.1 ``Inconclusive'' (an investigation could be justified to obtain them) and GPT-5-nano ``Inconclusive'' (the data are needed first).
Each model generated one period report, so the difference could come from the model or noise; for similar work, repeated generations are needed before these findings can be used as concrete evidence.



\subsection{Evidence Ablation}


Our results show that air quality data substantially influences model performance, but weather information and citizen complaints have minimal impact (see Section~\ref{sec:results} and Table~\ref{tab:ablation}). The limited effect of weather data is likely because its main purpose is to connect data streams through a puff travel model to select the most relevant air quality monitor, making it redundant as evidence in paragraphs. Additionally, while this ablation only removed weather details from event paragraphs, entirely removing weather data would break the pipeline by severing the link between events, monitors, and complaints, proving its structural necessity for data aggregation rather than direct evidence.
Although citizen complaints are critical for triggering follow-up inspections and belong in the overall workflow, they do not affect model performance significantly because environmental permits do not directly regulate them, leading the LLMs to fail understand the importance of this data stream. Instead, environmental service experts should determine the real effect of citizen complaints, which would allow LLMs to understand which metrics and evidence warrant an investigation.


\subsection{Design Implications for Involving Environmental Service Experts in the Loop}

Based on the discussion points in previous subsections, the following design implications for involving environmental service experts in the loop is established.

\emph{Determine which AQ monitors and chemical components should be used to act as accurate indicators of the pollution from emission sources}. As discussed in Evidence Aggregation, it is technically not possible to obtain AQ readings prior to a formal inspection. Thus, approximation of source emissions is a must.

\emph{Be involved in designing the reference model for aligning citizen complaints, air quality data, and atmospheric conditions.} We have used a simple puff model to align the data and select the air quality monitors that possibly had the relevant readings. Environmental service can provide expert knowledge to adjust and improve the model for correct alignment of data.

\emph{Clarify which permit condition is related to which facility.} Permit Retrieval has shown that this information is critical in reducing the errors of wrong citations. To clearly determine which facility is regulated by which permit, we need evaluation from experts.

\emph{Be involved in designing retrieval metrics to
fit the legal inspection workflow.} General retrieval and prompt designs may not reflect the model performance of a domain-specific task well, and thus, we need experts to be involved in designing domain-specific metrics and instructions.

\emph{Provide data to link investigation decisions and the generated report to post-train the LLM.} As shown in Report Generation, LLMs has uncertainties when providing recommendations even using the same event paragraphs. Thus, we need the investigation decisions data to post-train the LLMs to better align with the legal reasoning task.

\emph{Determine the role of citizen complaints in triggering a follow-up inspection so that LLMs can better use the information.} Currently, the
environmental permit does not directly regulate citizen complaints, and thus the LLM has no technical reason to use the information in it.

\subsection{Limitations}
\label{sec:limitations}

We describe the limitations of this work below.

\emph{Study Time Period.}
As a preliminary study to demonstrate feasibility and reveal ways to involve experts in the loop, we only evaluated the pipeline using data in March~2025. Future work should use more data, ideally covering one year.

\emph{Retrieval inputs.}
3 of the 42 conditions exceed the 512 tokens that the embedding models read, and the re-ranker reads the event paragraph and a condition in one window, leading to some parts of the permit being cut off.
The longest condition (1.2.20 H) is retrieved for 73\% of the events by Variant~B but for at most 3\% by Variants~C and~D, where its sub-clauses are retrieved instead.
Future work should inspect whether the information loss due to the length limit impact the performance.
Also, the cross-encoder, trained on English, was not validated on Dutch.

\emph{Generalisability.}
The insights concern two coke plants of one factory under Dutch permits, which means that the pipeline may not be generalisable to other contexts.
The plant filter relies on emission-point codes and facility names, and the pollutant groups reflect coke production, so another case needs these information re-derived.
\section{Conclusion}
\label{sec:conclusion}
We have built and evaluated a prototype that indicates the feasibility of using LLMs to support Dutch environmental law enforcement by presenting a retrieval-augmented generation (RAG) pipeline that connects public emission evidence to permit conditions for compliance screening.
Moreover, throughout the study, we have identified several design issues in each stage of the pipeline, reflected them with the experiment results, and suggest design implications about how to involve environmental service experts in the loop, which provides a road map for follow-up integration of this work into the local context.
Our goal for this work is not to replace experts in law enforcement, and rather, we envision a future that LLMs can be used and evaluated properly with an in-house infrastructure to support law enforcement, ultimately leading to timely response to citizen requests and better legal obligation checks.

\clearpage
\section*{Ethical Considerations Statement}

This study uses public data on industrial emission events, environmental
permits, air-quality measurements and weather, and complaints by residents, for
a historical month (March 2025). No human participants were involved and the
environmental service was not contacted. Information such as individual complaints are aggregated into counts and descriptors per postcode
area before they reach the language models and this paper.

The system is a screening aid and not a decision tool. Its outputs are not
compliance determinations: the period recommendation differed between models
and, under evidence removal, between repeats, and reports can cite conditions
that do not apply or miss ones that do. The risks considered are over-reliance
on the output (automation bias), attention shifted towards some areas or
companies by the design of a screening tool~\cite{benami2021}, complaint
evidence that reflects who reports as much as what happens, and a company being
implied to have violated its permit. The study limits these risks by treating
complaints as context and not as labels, by evaluating plant scope separately
from relevance, by reporting the noise of the measurements, and by not
publishing event-level outputs. Checking the plant
labels, the validity of the conditions and the reports with the environmental
service is planned as the next design iteration.

The language models were accessed through an institutional API
proxy, and only public data were sent to them.

\section*{Generative AI Usage Statement}

Generative AI was used in preparing this manuscript, code and analysis, as
follows. Claude (Anthropic; Claude Sonnet 5.5) was used
(i)~to suggest and refine passages of the abstract, introduction, related work,
methodology, results and discussion from the authors' code, result files, notes
and supervisor comments; the authors edited these passages and decided what to
include;
(ii)~to review the manuscript for inconsistencies, factual and grammatical
errors, and to suggest cuts, which the authors checked and applied;
(iii)~to compare the manuscript text with the analysis code and the data files
and to point out inconsistencies, which the authors checked and corrected;
(iv)~to recompute tables and statistics from the raw output files as a
cross-check of the authors' own runs;
(v)~to write and debug analysis and evaluation scripts (a corrected evaluation
notebook for the evidence-aggregation checks, a check of the token length of
the permit conditions, and changes to the evidence-removal script and its
tests), all of which the authors ran;
(vi)~to write English summaries of Dutch permit conditions for a label audit;

Human verification was constant, comparing generated suggestions using own data, results and findings. All text and code produced with the tool were reviewed, edited and, for code, run by the authors, and every number in the tables was checked against the output files.

The language models that are studied in the paper (GPT-5.1, GPT-5-mini,
GPT-5-nano and gpt-oss-120b) are components of the evaluated system and of its
evaluation, as described in the Methodology; they were not used to write the
paper.

All authors take responsibility for the correctness and originality of the
content, including references and citations.

\section*{Acknowledgements}
We thank the RIVM (National Institute for Public Health and Environment) and KNMI (Royal Netherlands Meteorological Institute) for making the air quality and weather data publicly available. We also thank the OD NZKG (North Sea Canal Area Environmental Service) for providing the citizen reports, the emission events list, and the environmental permit data. We also thank the UvA AI Chat for providing the LLM computing infrastructure.


\bibliographystyle{ACM-Reference-Format}
\bibliography{bibliographies/references}

@misc{han2025,
    title        = {{Evaluating LLM-based Approaches to Legal Citation
                     Prediction: Domain-specific Pre-training, Fine-tuning,
                     or RAG? A Benchmark and an Australian Law Case Study}},
    year         = {2025},
    author       = {Han, Jiuzhou and Burgess, Paul and Shareghi, Ehsan},
    eprint       = {2412.06272},
    archivePrefix = {arXiv},
    doi          = {10.48550/arXiv.2412.06272}
}

@article{buenopatin2026,
    title   = {{Jobs vs the environment: Struggles over the green steel
                plan at IJmuiden, The Netherlands}},
    year    = {2026},
    journal = {Environment and Planning E: Nature and Space},
    author  = {Bueno Patin, Antonio and Kaika, Maria},
    pages   = {25148486261440795},
    doi     = {10.1177/25148486261440795}
}

@misc{es2025ragas,
      title={Ragas: Automated Evaluation of Retrieval Augmented Generation}, 
      author={Shahul Es and Jithin James and Luis Espinosa-Anke and Steven Schockaert},
      year={2025},
      eprint={2309.15217},
      archivePrefix={arXiv},
      primaryClass={cs.CL},
      url={https://arxiv.org/abs/2309.15217}, 
}

@misc{gao2024ragsurvey,
    title        = {{Retrieval-Augmented Generation for Large Language
                     Models: A Survey}},
    year         = {2024},
    author       = {Gao, Yunfan and Xiong, Yun and Gao, Xinyu and
                    Jia, Kangxiang and Pan, Jinliu and Bi, Yuxi and
                    Dai, Yi and Sun, Jiawei and Wang, Meng and Wang, Haofen},
    eprint       = {2312.10997},
    archivePrefix = {arXiv},
    doi          = {10.48550/arXiv.2312.10997}
}

@inproceedings{guha2023legalbench,
    title        = {{LegalBench: A Collaboratively Built Benchmark for
                     Measuring Legal Reasoning in Large Language Models}},
    year         = {2023},
    booktitle    = {Advances in Neural Information Processing Systems
                     (NeurIPS)},
    volume       = {36},
    author       = {Guha, Neel and Nyarko, Julian and Ho, Daniel E. and
                    R{\'e}, Christopher and Chilton, Adam and
                    Narayana, Aditya and Chohlas-Wood, Alex and
                    Peters, Austin and Waldon, Brandon and
                    Rockmore, Daniel N. and Zambrano, Diego and
                    Talisman, Divya and Hoque, Enam and Surani, Faiz and
                    Fagan, Frank and Sarfaty, Galit and
                    Dickinson, Gregory M. and Porat, Haggai and
                    Hegland, Jason and Li, Zehua},
    eprint       = {2308.11462},
    archivePrefix = {arXiv},
    doi          = {10.48550/arXiv.2308.11462}
}

@article{hino2018ml,
  author  = {Hino, Miyuki and Benami, Elinor and Brooks, Nina},
  title   = {Machine Learning for Environmental Monitoring},
  journal = {Nature Sustainability},
  volume  = {1},
  number  = {10},
  pages   = {583--588},
  year    = {2018},
  doi     = {10.1038/s41893-018-0142-9}
}

@article{irwin1979,
    title   = {{A Theoretical Variation of the Wind Profile Power-Law
                Exponent as a Function of Surface Roughness and Stability}},
    year    = {1979},
    journal = {Atmospheric Environment},
    author  = {Irwin, J. S.},
    volume  = {13},
    number  = {1},
    pages   = {191--194},
    doi     = {10.1016/0004-6981(79)90260-9}
}

@inproceedings{lewis2021,
    title        = {{Retrieval-Augmented Generation for Knowledge-Intensive
                     NLP Tasks}},
    year         = {2020},
    booktitle    = {Advances in Neural Information Processing Systems
                     (NeurIPS)},
    volume       = {33},
    author       = {Lewis, Patrick and Perez, Ethan and Piktus, Aleksandra
                    and Petroni, Fabio and Karpukhin, Vladimir and
                    Goyal, Naman and K{\"u}ttler, Heinrich and
                    Lewis, Mike and Yih, Wen-tau and
                    Rockt{\"a}schel, Tim and Riedel, Sebastian and
                    Kiela, Douwe},
    eprint       = {2005.11401},
    archivePrefix = {arXiv},
    doi          = {10.48550/arXiv.2005.11401}
}

@misc{magesh2024,
    title  = {{Hallucination-Free? Assessing the Reliability of Leading
               AI Legal Research Tools}},
    year   = {2024},
    author = {Magesh, Varun and Surani, Faiz and Dahl, Matthew and
              Suzgun, Mirac and Manning, Christopher D. and Ho, Daniel E.}
}

@misc{meyur2025,
    title        = {{Benchmarking LLMs for Environmental Review
                     and Permitting}},
    year         = {2025},
    author       = {Meyur, Rounak and Phan, Hung and Hayashi, Koby and
                    Stewart, Ian and Sharma, Shivam and
                    Chaturvedi, Sarthak and Parker, Mike and
                    Nally, Dan and Montgomery, Sadie and
                    Pazdernik, Karl and Jannesari, Ali and
                    Halappanavar, Mahantesh and Munikoti, Sai and
                    Horawalavithana, Sameera and Acharya, Anurag},
    eprint       = {2407.07321},
    archivePrefix = {arXiv},
    doi          = {10.48550/arXiv.2407.07321}
}

@book{turner1994,
    title     = {{Workbook of Atmospheric Dispersion Estimates:
                  An Introduction to Dispersion Modeling}},
    year      = {1994},
    publisher = {Lewis Publishers},
    author    = {Turner, D. Bruce},
    edition   = {2nd},
    isbn      = {978-1566700239}
}

@inproceedings{zheng2023,
    title        = {{Judging LLM-as-a-Judge with MT-Bench
                     and Chatbot Arena}},
    year         = {2023},
    booktitle    = {Advances in Neural Information Processing Systems
                     (NeurIPS)},
    volume       = {36},
    pages        = {46595--46623},
    author       = {Zheng, Lianmin and Chiang, Wei-Lin and Sheng, Ying
                    and Zhuang, Siyuan and Wu, Zhanghao and
                    Zhuang, Yonghao and Lin, Zi and Li, Zhuohan and
                    Li, Dacheng and Xing, Eric P. and Zhang, Hao and
                    Gonzalez, Joseph E. and Stoica, Ion},
    eprint       = {2306.05685},
    archivePrefix = {arXiv},
    doi          = {10.48550/arXiv.2306.05685}
}

@article{ji2021citizen,
    title   = {{Spatial Analysis of Citizens' Environmental Complaints
                in China: Implications in Environmental Monitoring
                and Governance}},
    year    = {2021},
    journal = {International Journal of Environmental Research
               and Public Health},
    author  = {Ji, Xuepeng and Tong, Daoqin and Cheng, Lisha and
               Chuai, Xiaowei and Mao, Xiyan and Liu, Binglin and
               Huang, Xianjin},
    volume  = {18},
    number  = {18},
    pages   = {9674},
    doi     = {10.3390/ijerph18189674}
}

@techreport{epa2018,
    title       = {{Technical Assistance Document for the Reporting of
                    Daily Air Quality - the Air Quality Index (AQI)}},
    year        = {2018},
    institution = {United States Environmental Protection Agency},
    author      = {{US EPA}},
    url         = {https://www.airnow.gov/sites/default/files/2020-05/aqi-technical-assistance-document-sept2018.pdf}
}

@techreport{eubref2013,
    title       = {{Best Available Techniques (BAT) Reference Document
                    for Iron and Steel Production}},
    year        = {2013},
    institution = {European Commission, Joint Research Centre,
                   Institute for Prospective Technological Studies},
    author      = {Remus, Rainer and Aguado-Monsonet, Miguel A. and
                   Roudier, Serge and {Delgado Sancho}, Luis},
    doi         = {10.2791/97469},
    url         = {https://bureau-industrial-transformation.jrc.ec.europa.eu/sites/default/files/2019-11/IS_Adopted_03_2012.pdf}
}

@techreport{emepeea2023,
    title       = {{EMEP/EEA Air Pollutant Emission Inventory Guidebook
                    2023: 2.C.1 Iron and Steel Production}},
    year        = {2023},
    institution = {European Environment Agency},
    author      = {Jurich, Kristina and Kuenen, Jeroen and
                   Berdowski, Jan and {van der Most}, Pieter and
                   Slager, J. M. and Mulder, W. and
                   Hlawiczka, Stanislaw and Fudala, Janina and
                   Bloos, Jan Pieter and Verhoeve, P. and
                   Quass, Ulrich and Pierce, Martin and
                   Pulles, Tinus and Appelman, Wilfred and
                   Rentz, Otto and Karl, Ute and Woodfield, Mike},
    url         = {https://www.eea.europa.eu/en/analysis/publications/emep-eea-guidebook-2023/part-b-sectoral-guidance-chapters/2-industrial-processes-and-product-use/2-c-metal-production/2-c-1-iron-and/@@download/file},
}

@techreport{odnzkg2023,
    title       = {{Kooksgasfabrieken Tata Steel onder verscherpt toezicht}},
    year        = {2023},
    institution = {Omgevingsdienst Noordzeekanaalgebied},
    author      = {{OD NZKG}},
    url         = {https://odnzkg.nl/nieuws/kooksgasfabrieken-tata-steel-onder-verscherpt-toezicht/}
}

@techreport{odnzkg2024jaarbeeld,
    title       = {{OD NZKG Jaarbeeld 2024}},
    year        = {2025},
    institution = {Omgevingsdienst Noordzeekanaalgebied},
    author      = {{OD NZKG}},
    url         = {https://odnzkg.nl/wp-content/uploads/2025/09/OD-NZKG-jaarbeeld-2024.pdf},

}

@misc{rivm_luchtmeetnet,
    title        = {{RIVM Luchtmeetnet -- National Air Quality
                     Monitoring Network}},
    year         = {2025},
    author       = {{Rijksinstituut voor Volksgezondheid
                     en Milieu (RIVM)}},
    url          = {https://data.rivm.nl/data/luchtmeetnet/}
}

@misc{cbs_pc4,
    title        = {{CBS Postcode gebiedsindelingen --
                     Gegevens per postcode}},
    year         = {2024},
    author       = {{Centraal Bureau voor de Statistiek (CBS)}},
    url          = {https://www.cbs.nl/nl-nl/dossier/nederland-regionaal/geografische-data/gegevens-per-postcode}
}

@misc{knmi_uurgegevens,
    title        = {{Uurgegevens van het weer in Nederland}},
    year         = {2025},
    author       = {{Koninklijk Nederlands Meteorologisch
                     Instituut (KNMI)}},
    url          = {https://www.knmi.nl/nederland-nu/klimatologie/uurgegevens}
}

@misc{odnzkg_tatasteel,
    title        = {{Gisviewer Tata Steel}},
    year         = {2026},
    author       = {{Omgevingsdienst Noordzeekanaalgebied (OD NZKG)}},
    url          = {https://gisviewer.odnzkg.nl/index.php?@TataSteel}
}

@techreport{rivm2023tata,
  author      = {Geelen, L. M. J. and Bogers, R. P. and Elberse, J. E. and Houthuijs, D. and Montforts, M. H. M. M. and Schuijff, M. and Smetsers, R. C. G. M. and de Vries, A. and Wesseling, J. and Wijten, J. H. J.},
  title       = {De bijdrage van {Tata Steel Nederland} aan de gezondheidsrisico's van omwonenden en de kwaliteit van hun leefomgeving},
  institution = {Rijksinstituut voor Volksgezondheid en Milieu (RIVM)},
  number      = {2023-0171},
  year        = {2023},
  doi         = {10.21945/RIVM-2023-0171},
  note        = {English title: The contribution of Tata Steel to the health risks of local residents and the quality of their surroundings}
}

@misc{gsnh2023vth,
  author = {{Gedeputeerde Staten van Noord-Holland}},
  title  = {Informatievoorziening {VTH} provinciale bedrijven},
  year   = {2023},
  month  = feb,
  note   = {Letter to Provinciale Staten, ref. 1951630/1972663},
  url    = {https://noord-holland.nl/bestanden/pdf/Brief%20aan%20PS%20over%20Informatievoorziening%20VTH%20provinciale%20bedrijven.pdf}
}

@misc{gsnh2025bezwaar,
  author = {{Gedeputeerde Staten van Noord-Holland}},
  title  = {Beslissing op bezwaar Lasten onder Dwangsom overschrijden emissiegrenswaarden {KGF} 1 en 2 {Tata Steel}},
  year   = {2025},
  month  = nov,
  note   = {Letter to Provinciale Staten},
  url    = {https://www.noord-holland.nl/bestanden/pdf/Tata_Steel/Brief%20aan%20PS%20over%20beslissing%20op%20bezwaar%20Lasten%20onder%20Dwangsom%20overschrijden%20emissiegrenswaarden%20KGF%201%20en%202%20Tata%20Steel.pdf}
}

@misc{gsnh2026intrekking,
  author = {{Gedeputeerde Staten van Noord-Holland}},
  title  = {Intrekken vergunningen Kooks- en Gasfabriek 1 en 2 {Tata Steel}},
  year   = {2026},
  month  = jun,
  note   = {Letter to Provinciale Staten, 11 June 2026},
  url    = {https://www.noord-holland.nl/bestanden/pdf/Gezonde_leefomgeving_Milieu/Brief%20aan%20PS%20over%20Intrekken%20vergunningen%20Kooks-%20en%20Gasfabriek%201%20en%202%20Tata%20Steel%20inclusief%20afdoeningsvoorstel%20motie%20Ml%206-2024.pdf}
}

@misc{odnzkg2026overons,
  author       = {{Omgevingsdienst Noordzeekanaalgebied}},
  title        = {Over ons},
  year         = {2026},
  url          = {https://odnzkg.nl/over-ons/},
  lastaccessed = {October 16, 2026}
}

@techreport{vanaartsen2021,
  author      = {{Adviescommissie Vergunningverlening, Toezicht en Handhaving}},
  title       = {Om de leefomgeving: Omgevingsdiensten als gangmaker voor het bestuur},
  institution = {Adviescommissie Vergunningverlening, Toezicht en Handhaving},
  year        = {2021},
  month       = mar,
  note        = {Commissioned by the State Secretary for Infrastructure and Water Management; committee chaired by J.J. van Aartsen; 4 March 2021},
  url         = {https://zoek.officielebekendmakingen.nl/blg-972065.pdf}
}

@inproceedings{benami2021,
  author    = {Benami, Elinor and Whitaker, Reid and La, Vincent and Lin, Hongjin and Anderson, Brandon R. and Ho, Daniel E.},
  title     = {The Distributive Effects of Risk Prediction in Environmental Compliance: Algorithmic Design, Environmental Justice, and Public Policy},
  booktitle = {Proceedings of the 2021 ACM Conference on Fairness, Accountability, and Transparency},
  series    = {FAccT '21},
  location  = {Virtual Event, Canada},
  publisher = {Association for Computing Machinery},
  address   = {New York, NY, USA},
  pages     = {90--105},
  year      = {2021},
  doi       = {10.1145/3442188.3445873}
}

@inproceedings{cormack2009,
  author    = {Cormack, Gordon V. and Clarke, Charles L. A. and Buettcher, Stefan},
  title     = {Reciprocal Rank Fusion Outperforms Condorcet and Individual Rank Learning Methods},
  booktitle = {Proceedings of the 32nd International ACM SIGIR Conference on Research and Development in Information Retrieval},
  year = {2009}, doi = {10.1145/1571941.1572114}
}

@misc{wang2024me5,
  author = {Wang, Liang and Yang, Nan and Huang, Xiaolong and Yang, Linjun and Majumder, Rangan and Wei, Furu},
  title = {Multilingual E5 Text Embeddings: A Technical Report},
  year = {2024}, eprint = {2402.05672}, archivePrefix = {arXiv}
}

@inproceedings{Akpinar_2021, series={FAccT ’21},
   title={The effect of differential victim crime reporting on predictive policing systems},
   url={http://dx.doi.org/10.1145/3442188.3445877},
   DOI={10.1145/3442188.3445877},
   booktitle={Proceedings of the 2021 ACM Conference on Fairness, Accountability, and Transparency},
   publisher={ACM},
   author={Akpinar, Nil-Jana and De-Arteaga, Maria and Chouldechova, Alexandra},
   year={2021},
   month=Mar, pages={838–849},
   collection={FAccT ’21} 
}

@misc{nogueira2020passagererankingbert,
      title={Passage Re-ranking with BERT}, 
      author={Rodrigo Nogueira and Kyunghyun Cho},
      year={2019},
      eprint={1901.04085},
      archivePrefix={arXiv},
      primaryClass={cs.IR},
      url={https://arxiv.org/abs/1901.04085}, 
}

@techreport{epa1995isc3v2,
  author      = {{U.S. Environmental Protection Agency}},
  title       = {User's Guide for the Industrial Source Complex ({ISC3}) Dispersion Models, Volume {II}: Description of Model Algorithms},
  institution = {U.S. Environmental Protection Agency, Office of Air Quality Planning and Standards},
  address     = {Research Triangle Park, NC},
  type        = {Technical Report},
  number      = {EPA-454/B-95-003b},
  year        = {1995},
  month       = sep,
  url         = {https://gaftp.epa.gov/aqmg/SCRAM/models/other/isc3/isc3v2.pdf},
  note        = {Accessed November 2026}
}

@article{robertson2009bm25,
  author  = {Robertson, Stephen and Zaragoza, Hugo},
  title   = {The Probabilistic Relevance Framework: {BM25} and Beyond},
  journal = {Foundations and Trends in Information Retrieval},
  volume  = {3},
  number  = {4},
  pages   = {333--389},
  year    = {2009},
  doi     = {10.1561/1500000019}
}

\clearpage

\appendix
\begin{appendices}


\section{March compared with the other months}
\label{app:months}

\begin{table}[h]
\caption{Events per month in 2025. AQ coverage is the share of events with an AQ severity. Complaint events have at least one matched complaint. Matched pairs count complaint--event matches; one complaint can match more than one event.}
\small
\begin{tabular}{@{}lrrrrrr@{}}
\toprule
Month & Events & Plant 1 & Plant 2 & AQ coverage (\%) & Complaint events & Matched pairs \\
\midrule
Jan & 27 & 18 & 9 & 70 & 2 & 3 \\
Feb & 41 & 28 & 13 & 88 & 1 & 1 \\
\textbf{Mar} & 74 & 27 & 47 & 86 & 10 & 32 \\
Apr & 72 & 32 & 40 & 86 & 4 & 4 \\
May & 67 & 32 & 35 & 85 & 10 & 13 \\
Jun & 54 & 22 & 32 & 89 & 4 & 4 \\
Jul & 79 & 44 & 35 & 77 & 8 & 8 \\
Aug & 65 & 34 & 31 & 80 & 3 & 4 \\
Sep & 56 & 28 & 28 & 89 & 1 & 1 \\
Oct & 63 & 27 & 36 & 89 & 5 & 7 \\
Nov & 56 & 33 & 23 & 75 & 4 & 4 \\
Dec & 53 & 33 & 20 & 75 & 8 & 12 \\
\midrule
Year & 707 & 358 & 349 & 83 & 60 & 93 \\
\bottomrule
\end{tabular}
\end{table}

\section{Plume cone constants and margin}
\label{app:margin}

The half-angle $\theta_s(x) = c_s - d_s \ln x$ (degrees, $x$ in km, used from 0.1~km) follows the Pasquill--Gifford curves in the angle form of the ISC3 model (rural). The constants are:

\begin{table}[h]
\small
\begin{tabular}{@{}lcccccc@{}}
\toprule
Class & A & B & C & D & E & F \\
\midrule
$c_s$ & 24.167 & 18.333 & 12.500 & 8.333 & 6.250 & 4.1667 \\
$d_s$ & 2.5334 & 1.8096 & 1.0857 & 0.72382 & 0.54287 & 0.36191 \\
\bottomrule
\end{tabular}
\end{table}

For the intermediate classes AB, BC and CD, the constants of the two neighbouring classes are averaged; this convention has not been checked in the primary source.

\paragraph{Stability class and wind speed.}
By day (07:00 to 19:00 with solar radiation above 10~J/cm$^2$), the Pasquill--Gifford class is taken from solar radiation and wind speed; otherwise it is taken from cloud cover and wind speed, using the table of Turner~\cite{turner1994}, and class D is used if no row applies.
The wind speed is scaled from 10~m to 35~m with the exponent $\alpha$ of Irwin~\cite{irwin1979}: 0.08 (A), 0.09 (B), 0.11 (C), 0.16 (D), 0.32 (E) and 0.54 (F), and the mean of the two exponents for AB, BC and CD. 
The Turner~\cite{turner1994} daytime thresholds in W/m$^2$ are converted to KNMI global radiation thresholds in J/cm$^2$/h using the conversion factor 1~J/cm$^2$/h = 2.778~W/m$^2$, yielding the values 216, 108, and 54~J/cm$^2$/h.

On the 11 non-March months of 2025, air quality (AQ) severity at the six monitors, minus each monitor's own baseline when not downwind, stays at 0.7--1.1 up to 20 degrees beyond the 10\% plume edge and falls to about 0.5 further out (difference 0.34, bootstrap 95\% CI 0.20--0.49). Rounding alone from the KNMI wind direction resolution (5 degrees) represents a strict lower bound that removed real signal, dropping 67 events with an in-cone monitor and 22 matched citizen reports. However, several caveats apply to this calibration: the excess signal away from the plume cone may partly reflect stagnant-air pollution; a single margin is applied across all stability classes (where stable-class bins hold only about 23 pairs); and the footprint is measured directly at monitors rather than across postcode areas.

\section{Monitors, complaint descriptors and Dutch terms}
\label{app:dutch}

\begin{table}[h]
\caption{AQ monitors and their distance to the chimney of each plant. Gas monitors also measure gases; the other monitors measure only PM$_{10}$ and PM$_{2.5}$.}
\small
\begin{tabular}{@{}lrrl@{}}
\toprule
Station ID & Plant 1 (km) & Plant 2 (km) & Measures \\
\midrule
NL49551 & 1.39 & 1.97 & gases and PM \\
NL49553 & 2.29 & 2.36 & gases and PM \\
NL49557 & 1.95 & 1.88 & PM only \\
NL49570 & 2.60 & 4.16 & PM only \\
NL49572 & 1.16 & 3.11 & PM only \\
NL49573 & 2.25 & 0.32 & PM only \\
\bottomrule
\end{tabular}
\end{table}

\begin{table}[h]
\caption{Complaint descriptors (Dutch original and English translation) by group.}
\small
\begin{tabular}{@{}lll@{}}
\toprule
Group & Dutch & English \\
\midrule
Coke-related & Kooks & coke \\
 & Zwavel/rotte eieren & sulphur / rotten eggs \\
 & Bakeliet/Fenol & bakelite / phenol \\
 & Teer & tar \\
 & Ammoniak & ammonia \\
Particulate & Zwart/grijs & black / grey \\
 & Stofwolk & dust cloud \\
Other & all other descriptors & \\
\bottomrule
\end{tabular}
\end{table}

\section{Prompts}
\label{app:prompts}

\subsection{LLM Roles}
\paragraph{System role}
\begin{lstlisting}[style=prompt]
You are a compliance assessment assistant for OD NZKG. You may only cite permit conditions and evidence present in the provided context. Do not infer rules not explicitly retrieved. If a relevant permit condition is absent from context, state this explicitly. All output must be written in English regardless of the language of the source documents.
\end{lstlisting}

\paragraph{Hallucination guard}
\begin{lstlisting}[style=prompt]
After generating your response, verify that every voorschriftnummer you cite exists in the retrieved context above. If any cited ID is not present in context, remove it and note the omission.
\end{lstlisting}

\subsection{Event Report Prompt}
\begin{lstlisting}[style=prompt]
=== TASK ===
Write a concise compliance mini-report for this single event. Structure it as:
1. Event summary (1-2 sentences)
2. Severity assessment (reference aq_severity_z, citizen descriptor counts, offshore_flag)
3. Permit conditions potentially relevant to this event (cite each by voorschriftnummer)
4. Preliminary compliance observations (what may or may not align with retrieved permits)
5. Data gaps (any missing evidence or absent permit conditions needed for full assessment)
Respond entirely in English.
\end{lstlisting}

\subsection{Period Report Prompt}
\begin{lstlisting}[style=prompt]
=== TASK ===
Based on the per-event reports above, write a consolidated compliance assessment for this period. Structure it as:
1. Executive summary (3-5 sentences)
2. Recurring permit conditions of concern (group events by voorschriftnummer)
3. Highest-severity events (rank by aq_severity_z, note any with citizen descriptor evidence)
4. Overall recommendation: does this period warrant a formal investigation and why?
   Begin your answer to item 4 with one of: Yes / No / Inconclusive.
5. Data limitations for this assessment

Only cite voorschriftnummers that appear in the per-event reports above.
Write the full assessment in English.
\end{lstlisting}

\subsection{LLM-Judge Rubric}
\label{app:rubric}
The rubric below was given to the judge (gpt-oss-120b) together with the event
paragraph, the retrieved permit conditions and the report to be evaluated.
``Chimney'' in the rubric refers to the coke plant (KGF1 or KGF2).

\begin{lstlisting}[style=prompt]
Score each dimension 1, 2, or 3:
  1 = Absent or clearly incorrect
  2 = Partially present or incomplete
  3 = Fully present and correct

DIMENSION 1 - legal_grounding
Does every compliance claim in sections 3 and 4 cite a specific voorschrift
that exists in the RETRIEVED PERMIT CONDITIONS provided below?
Penalise if:
  - a voorschrift is cited that is NOT in the retrieved set
  - a voorschrift is cited that applies to the WRONG chimney
    (e.g. citing a KGF2-specific condition for a KGF1 event)
  - compliance conclusions are stated without any permit citation

DIMENSION 2 - evidence_integration
Are the AQ sensor readings, atmospheric conditions (wind, PG class),
and citizen report data used to support or qualify the legal observations
in sections 3 and 4? Or are they only listed in section 2 without connecting
to the permit reasoning?
Penalise if:
  - evidence is listed in section 2 but never referenced in sections 3-4
  - permit citations appear with no connection to the observed evidence
  - the report treats evidence and permit conditions as two separate topics

DIMENSION 3 - epistemic_calibration
Does the report correctly acknowledge that ambient sensor readings (ug/m3
at 1-4 km distance) cannot be directly compared to stack emission limits
(mg/Nm3 at source)? Does it hedge appropriately rather than claiming
compliance or non-compliance from ambient data alone?
Penalise if:
  - the report claims a permit limit is exceeded based only on ambient data
  - the report claims compliance is demonstrated from ambient data
  - the metric gap is not mentioned at all

DIMENSION 4 - data_gap_identification
Does section 5 correctly identify the specific evidence that would be
needed to make a formal compliance determination (e.g. stack measurements,
continuous emission monitoring data, enforcement compound measurements)?
Penalise if:
  - data gaps are vague ("more information needed") without specifying what
  - gaps relevant to the specific retrieved voorschriften are missed
  - the section is absent or only one line
\end{lstlisting}

\end{appendices}

\end{document}